\documentclass[sigconf]{acmart}
\AtBeginDocument{%
  }
\usepackage{multirow}
\acmConference[ISMSI '25]{April 26--28,
  2025}{Tokyo, Japan}

\renewcommand\footnotetextcopyrightpermission[1]{}

\begin{document}

\title{Enhancing Link Prediction with Fuzzy Graph Attention Networks and Dynamic Negative Sampling}


\author{Jinming Xing}
\affiliation{%
  \institution{North Carolina State University}
  \country{}}
\email{jxing6@ncsu.edu}

\author{Ruilin Xing}
\affiliation{%
  \institution{Guangxi University}
  \country{}}
\email{ruilinxing8@gmail.com}


\author{Chang Xue}
\affiliation{%
  \institution{Yeshiva University}
  \country{}}
\email{cxue@mail.yu.edu}

\author{Dongwen Luo}
\affiliation{%
  \institution{South China University of Technology}
  \country{}}
\email{976267567ldw@gmail.com}

\renewcommand{\shortauthors}{Jinming et al.}

\begin{abstract}
  Link prediction is crucial for understanding complex networks but traditional Graph Neural Networks (GNNs) often rely on random negative sampling, leading to suboptimal performance. This paper introduces Fuzzy Graph Attention Networks (FGAT), a novel approach integrating fuzzy rough sets for dynamic negative sampling and enhanced node feature aggregation. Fuzzy Negative Sampling (FNS) systematically selects high-quality negative edges based on fuzzy similarities, improving training efficiency. FGAT layer incorporates fuzzy rough set principles, enabling robust and discriminative node representations. Experiments on two research collaboration networks demonstrate FGAT's superior link prediction accuracy, outperforming state-of-the-art baselines by leveraging the power of fuzzy rough sets for effective negative sampling and node feature learning.
\end{abstract}



\maketitle

\section{Introduction}
Link prediction has emerged as a crucial task in network analysis with extensive applications across diverse domains. In medical sciences, it aids in predicting protein-protein interactions and drug-target associations; in financial systems, it helps detect fraudulent transactions and assess credit risks; and in chemistry, it facilitates the discovery of novel molecular structures and chemical reactions. The ability to accurately predict potential connections in these complex networks has significant implications for scientific advancement and practical applications.

Graph Neural Networks (GNNs) have demonstrated remarkable success in link prediction tasks, primarily due to their inherent capability to capture and process structural information in graph-structured data. However, a critical limitation in existing GNN-based approaches lies in their negative sampling methodology. Contemporary methods typically employ random sampling strategies to select negative edges, disregarding the rich semantic and structural information encoded in node representations. This oversight significantly hampers the training process, resulting in slower convergence rates and suboptimal model performance. An ideal negative sampling mechanism should not only leverage node embeddings effectively but also adaptively select high-quality negative samples based on the model's current state, ensuring both dynamic responsiveness and sampling accuracy.

While various methodologies have been explored to enhance link prediction accuracy, the potential of fuzzy rough sets—a mathematical framework for measuring fuzzy relations and handling uncertainty—remains largely unexplored in the context of GNNs and link prediction. This theoretical framework offers unique advantages in capturing imprecise relationships and handling ambiguous data structures, making it particularly suitable for network analysis tasks.

To address these limitations and leverage the untapped potential of fuzzy rough sets, we propose a novel fuzzy rough sets-based negative sampling strategy called Fuzzy Negative Sampling (FNS). This approach systematically evaluates candidate negative edges through their fuzzy lower approximation values, selecting the top K candidates as negative training instances. Furthermore, we introduce Fuzzy Graph Attention Network (FGAT), an enhanced graph neural architecture designed to aggregate neighboring node information in a more robust and effective manner.

The main contributions of this work can be summarized as follows:
\begin{itemize}
    \item We introduce FNS, a novel negative sampling framework that leverages fuzzy rough sets theory to identify high-quality negative edges, significantly improving the effectiveness of the training process in link prediction tasks.
    \item We propose FGAT, an innovative graph attention network that incorporates fuzzy rough set principles to achieve more robust and discriminative node representations.
    \item We conduct comprehensive experiments across two real-world datasets, demonstrating the effectiveness of our proposed framework.
\end{itemize}

\section{Related Work}
\textbf{Graph Neural Networks} have demonstrated remarkable versatility across various graph-based learning tasks. In node classification, seminal works like GraphSAGE \cite{hamilton2017inductive} and Graph Attention Networks (GAT) \cite{velivckovic2017graph} have established foundational approaches for learning node representations through neighborhood aggregation. GCN \cite{kipf2016semi} introduced convolutional operations on graphs, enabling efficient feature propagation across network structures. For graph classification tasks, hierarchical pooling mechanisms have been developed, with DiffPool \cite{ying2018hierarchical} and TopKPool \cite{diehl2019edge} proposing learnable strategies to generate graph-level representations. In the context of link prediction, SEAL \cite{zhang2018link} pioneered the use of local subgraphs for edge existence prediction, while VGAE \cite{kipf2016variational} employed variational autoencoders for learning edge formation patterns. Recent advances include NGNN \cite{zhang2021nested}, which introduces neural architecture improvements specifically designed for link prediction tasks.

\textbf{Fuzzy rough sets theory}, initially proposed by Dubois and Prade \cite{dubois1990rough}, has evolved into a powerful framework for handling uncertainty and imprecision in data analysis. In feature selection, Jensen and Shen \cite{jensen2004fuzzy} developed fuzzy-rough attribute reduction algorithms that significantly outperform traditional approaches in identifying relevant features while maintaining information fidelity. The application of fuzzy rough sets in medical diagnosis has been exemplified by works such as \cite{xing2022weighted}, where they effectively handle the inherent uncertainty in patient data for more accurate disease classification. For uncertainty measurement, the framework has been extensively studied in theoretical works by \cite{gao2022parameterized} and \cite{ye2021novel}, establishing mathematical foundations for quantifying various types of uncertainty in data relationships. Recent developments include hybrid approaches combining fuzzy rough sets with deep learning and applications in big data analytics \cite{ji2021fuzzy} , demonstrating the framework's adaptability to modern computational challenges.

Except the innovative applications of Graph Neural Networks and fuzzy rough set theory in handling complex data and multi-task learning, against the backdrop of rapid advancements in machine learning, applications like knowledge graph technology in intelligent question-answering systems have become a key area of development for graph learning methods, effectively integrating multiple data sources to support flexible knowledge processing \cite{lin2021research}. Similarly, multi-model integration technology applied to automated generation systems has significantly enhanced content generation flexibility and quality, providing valuable insights for representation learning based on graph data \cite{yang2021research}. In network security, Graph Neural Networks have demonstrated strong generalization capabilities in supporting Botnet detection through machine learning, precisely identifying abnormal behaviors and enhancing network defense levels \cite{yang2022botnet}.

In recent years, the scope of machine learning applications has continuously expanded \cite{cheng24patch,cheng25unifying,cheng22deep,cheng22estimation,xing24traffic,cheng23shapnn}, especially in image processing and VR fields. Blockchain-enhanced image retrieval systems, for example, have brought revolutionary improvements in data security and retrieval efficiency \cite{zhao2022image}. In VR and robotic interaction, AI-vision-powered intelligent systems have explored new methods for balancing real-world interaction and virtual immersion, making human-computer interaction more natural and seamless \cite{yang2023balancing}. Furthermore, the integration of fuzzy rough set theory with deep learning has also achieved breakthroughs in traffic data prediction, enabling more accurate short-term forecasting through multi-source data fusion, effectively adapting to the dynamic demands of complex data environments \cite{deng2021short}. Overall, these technological innovations not only showcase the broad applicability of graph learning methods and fuzzy rough sets across different task scenarios but also reinforce their theoretical and practical value in big data and intelligent applications.

\section{Methodology}
\begin{figure}
    \centering
    \includegraphics[width=0.99\linewidth]{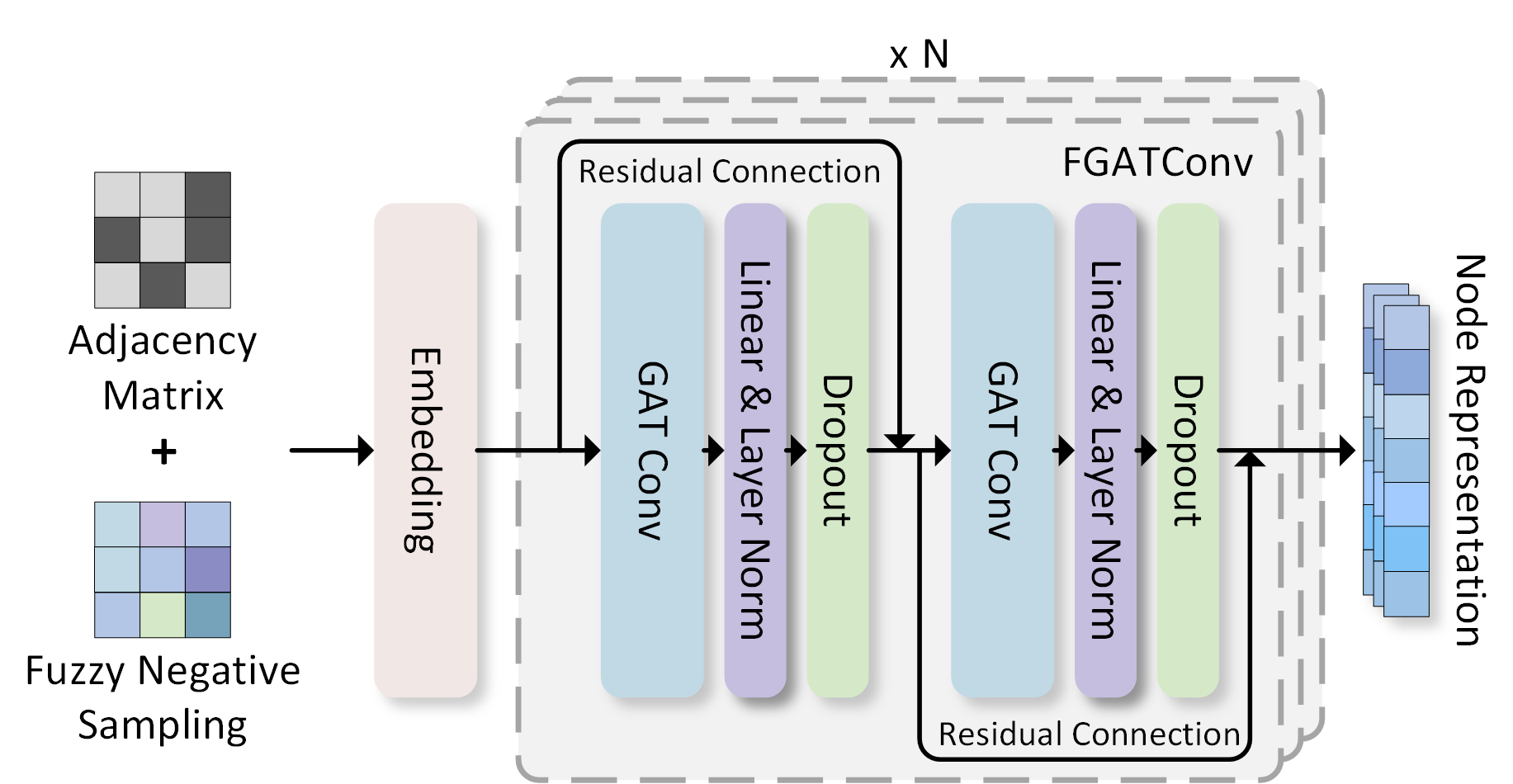}
    \caption{The FGAT Framework}
    \label{fig:framework}
\end{figure}

Figure \ref{fig:framework} illustrates our proposed framework, which consists of two main components:
\begin{itemize}
    \item \textbf{Fuzzy Negative Sampling:} A mechanism that selects high-quality negative edges based on fuzzy similarities, where negative edges with high fuzzy similarity are dynamically selected for the FGAT framework's training.
    \item \textbf{FGAT Convolution Layer:} A specially designed layer for effective neighbor node information aggregation. Multiple FGAT convolution layers are stacked to capture multi-hop information.
\end{itemize}

In the following sections, we first detail the computation of fuzzy similarities using fuzzy rough sets for high-quality negative edge selection. Subsequently, we elaborate on the FGAT architecture, followed by a comprehensive framework summary.

\subsection{Fuzzy Negative Sampling}
A fuzzy information system is defined as a tuple $(U,A,V,f)$, where $U$ represents a non-empty finite set of samples, $A$ denotes the finite set of sample attributes, $V$ represents the domain of all attributes in $A$, expressed as $V=\bigcup_i V_i$ where $V_i$ is the domain of attribute $i$, and $f$ is a mapping function $U\times A\rightarrow V$ \cite{xing2022weighted}.

For an attribute set $B\subseteq A$ and a fuzzy equivalence relation $R$, we can compute a coverage of the universe $U$. For a sample $x$, we denote its coverage under the fuzzy equivalence relation $R$ as $[x]_R$. The membership of a sample $y$ to the coverage $[x]_R$ is defined as $[x]_R(y)=R(x,y)$, where $R(x,y)$ quantifies the similarity between samples $x$ and $y$ under relation $R$. For any sample $x\in U$ and subset $X\subseteq A$, the fuzzy lower and upper approximations of sample $x$ to $X$ are defined as \cite{xing2022weighted}:
\begin{equation}
    \begin{split}
        \underline{R_S}X(x)&=\inf_{y\in U}S(N(R(x,y)),X(y)),\\
        \overline{R_T}X(x)&=\sup_{y\in U}T(R(x,y),X(y))
    \end{split}
    \label{eq:fuzzy approximations}
\end{equation}
where $S$ and $T$ represent fuzzy triangular conorm (S-norm) and fuzzy triangular norm (T-norm) respectively, and $N(x)=1-x$.

Using the conventional min-max version of $T$ and $S$ norms, for a set of samples $d_i$ of class $i$ and corresponding attribute set $B\subseteq A$, Equation \ref{eq:fuzzy approximations} can be reformulated as:
\begin{equation}
    \begin{split}
        \underline{R_B}d_i(x)&=\inf_{y\in U}\max(1-R(x,y),d_i(y)),\\
        \overline{R_B}d_i(x)&=\sup_{y\in U}\min(R(x,y),d_i(y))
    \end{split}
    \label{eq:fuzzy approximations min-max}
\end{equation}

To capture non-linear high-level similarities, $R$ typically employs kernel functions, including the Gaussian kernel: $k_G(x,y)=\exp(-\frac{||x-y||^2}{\delta})$, exponential kernel: $k_E(x,y)=\exp(-\frac{||x-y||}{\delta})$, and rational quadratic kernel: $k_R(x,y)=1-\frac{||x-y||^2}{||x-y||^2+\delta}$.

During each training epoch, negative links are dynamically selected based on their quality scores. For any potential negative link with end nodes $(x,y)$, the quality score is computed as:
\begin{equation}
    Score(x,y)=\alpha\times\underline{R_B}d_y(x)+(1-\alpha)\times\underline{R_B}d_x(y)
\end{equation}
where $\alpha$ is a hyperparameter.

While computing quality scores for all possible negative edges and selecting the top k would be optimal, this approach becomes computationally intractable for large dense graphs. For a graph with $N$ nodes and $E$ edges, there exist $N\times (N-1)-E$ potential directed negative edges. To address this computational challenge, we randomly select $2E$ negative edges and select the top $E$ edges among them. This strategy reduces the computational complexity from $N\times (N-1)-E$ to $2E$ while maintaining near-optimal performance.

The selected top $E$ negative edges are combined with the original positive edges to form the training dataset. To prevent class imbalance issues, we maintain an equal number of selected negative edges and original positive edges.

\subsection{FGAT Convolution Layer}
The FGAT convolution layer integrates GAT convolution layers with linear layers, incorporating layer normalization for training acceleration and dropout mechanisms for effective regularization.

Given an undirected graph $G = (V, E)$, where $V$ represents the set of nodes and $E$ denotes the set of edges, each node $v \in V$ is associated with a feature vector $\mathbf{h}_v \in \mathbb{R}^F$, where $F$ represents the dimension of input features per node. The FGAT layer aims to compute updated node representations $\mathbf{h}_v' \in \mathbb{R}^{F'}$, where $F'$ denotes the output feature dimension, by performing weighted aggregation of features from each node's neighborhood.

For a node pair consisting of node $v$ and its neighbor $u$, the attention coefficient $e_{vu}$ is computed through:
\begin{equation}
    e_{vu} = \text{LeakyReLU} \left( \mathbf{a}^T \left[ \mathbf{W} \mathbf{h}_v \parallel \mathbf{W} \mathbf{h}_u \right] \right)
\end{equation}
where:
\begin{itemize}
    \item $\mathbf{W} \in \mathbb{R}^{F' \times F}$ represents a learnable weight matrix that transforms node features linearly.
    \item $\parallel$ indicates vector concatenation.
    \item $\mathbf{a} \in \mathbb{R}^{2F'}$ denotes a learnable weight vector.
    \item LeakyReLU serves as the activation function, typically configured with a small negative slope (e.g., 0.2).
\end{itemize}

The attention coefficients then undergo normalization across each node's neighborhood using the softmax function:
\begin{equation}
    \alpha_{vu} = \frac{\exp(e_{vu})}{\sum_{k \in \mathcal{N}(v)} \exp(e_{vk})}
\end{equation}
where $\mathcal{N}(v)$ represents the neighborhood set of node $v$.

The normalized attention scores $\alpha_{vu}$ facilitate the computation of updated node features $\mathbf{h}_v'$ through weighted aggregation:
\begin{equation}
    \mathbf{h}_v' = \sigma \left( \sum_{u \in \mathcal{N}(v)} \alpha_{vu} \mathbf{W} \mathbf{h}_u \right)
\end{equation}
where $\sigma$ represents a non-linear activation function, typically implemented as ReLU.

To enhance model robustness and representational capacity, the GAT layers employ multi-head attention mechanisms. Specifically, K independent attention heads operate in parallel, each generating distinct attention coefficients and feature representations. These representations are subsequently concatenated to produce the final output:
\begin{equation}
    \mathbf{h}_v' = \parallel_{k=1}^K \sigma \left( \sum_{u \in \mathcal{N}(v)} \alpha_{vu}^{(k)} \mathbf{W}^{(k)} \mathbf{h}_u \right)
\end{equation}
where $\alpha_{vu}^{(k)}$ and $\mathbf{W}^{(k)}$ correspond to the attention coefficient and weight matrix of the k-th attention head, respectively.

Layer normalization \cite{xiong2020layer} is incorporated to stabilize and expedite the training process by normalizing layer inputs. For an input vector $\mathbf{h} = [h_1, h_2, \dots, h_d]$ with $d$ features, the normalized output $\mathbf{\hat{h}} = [\hat{h}_1, \hat{h}_2, \dots, \hat{h}_d]$ is computed as:
\begin{equation}
    \hat{h}_i = \frac{h_i - \mu}{\sqrt{\sigma^2 + \epsilon}}
\end{equation}
where $\mu = \frac{1}{d} \sum_{i=1}^d h_i$ represents the mean, $\sigma^2 = \frac{1}{d} \sum_{i=1}^d (h_i - \mu)^2$ denotes the variance, and $\epsilon$ is a small constant ensuring numerical stability. The final output $y$ is obtained through the application of learnable scaling parameter $\gamma$ and bias term $\beta$:
\begin{equation}
    y_i = \gamma \hat{h}_i + \beta
\end{equation}

\subsection{The FGAT Framework}
As illustrated in Figure \ref{fig:framework}, the FGAT framework operates through a systematic process that begins with dynamic negative edge selection during each training epoch, utilizing the given adjacency matrix. These dynamically selected negative edges, along with the existing positive edges, are subsequently processed by the FGAT layer in conjunction with their corresponding node embeddings. To effectively capture long-range dependencies within the graph structure, multiple FGAT layers are cascaded, with residual connections implemented to enhance training stability and information flow. Following the iterative processing through these layers, we obtain updated node representations $H=\{h_1,h_2,\dots,h_N\}$. The probability of link existence between any pair of nodes $x$ and $y$ is then computed as:
\begin{equation}
    P_r^{link}(x,y)=\text{Sigmoid}(h_xh_y^T)
\end{equation}

The framework's effectiveness stems from two key components: the fuzzy negative sampling technique, which efficiently identifies and selects high-quality negative edges, and the FGAT layer architecture, which performs iterative neighbor information aggregation. The empirical validation of this framework's performance is documented in the experiments section, demonstrating its effectiveness in link prediction tasks.
\begin{table*}[htbp]
    \centering
    \caption{Experiment Results}
    \begin{tabular}{lcccccccc}
        \toprule
        \multirow{2}[2]{*}{Methods} & \multicolumn{4}{c}{Ca-netscience} & \multicolumn{4}{c}{Ca-sandi-auths}                                                                                                                                                                    \\
        \cmidrule(lr){2-5} \cmidrule(lr){6-9}
                                    & \multicolumn{1}{c}{Precision}     & \multicolumn{1}{c}{Recall}         & \multicolumn{1}{c}{F1} & \multicolumn{1}{c}{ROC} & \multicolumn{1}{c}{Precision} & \multicolumn{1}{c}{Recall} & \multicolumn{1}{c}{F1} & \multicolumn{1}{c}{ROC} \\
        \midrule
        MLP                         & 0.4931                            & 0.5879                             & 0.5363                 & 0.5044                  & 0.5581                        & \textbf{1.0000}            & 0.7164                 & 0.6181                  \\
        GCN                         & 0.5903                            & 0.7363                             & 0.6553                 & 0.7078                  & 0.5417                        & 0.5417                     & 0.5417                 & 0.5304                  \\
        GraphSAGE                   & 0.5502                            & \textbf{0.8132}                    & 0.6563                 & 0.6277                  & 0.4651                        & 0.8333                     & 0.5970                 & 0.5885                  \\
        GAT                         & 0.6034                            & 0.7692                             & \textbf{0.6763}        & 0.6916                  & 0.6053                        & 0.9583                     & 0.7419                 & \textbf{0.7240}         \\
        FGAT                        & \textbf{0.6667}                   & 0.6593                             & 0.6630                 & \textbf{0.7422}         & \textbf{0.6216}               & 0.9583                     & \textbf{0.7541}        & 0.7170                  \\
        \bottomrule
    \end{tabular}%
    \label{tab:experiment results}%
\end{table*}%

\section{Experiments}
We conduct comparative evaluations of FGAT against several state-of-the-art baselines using two research collaboration network datasets. The experimental results demonstrate the superior performance of FGAT. In this section, we present detailed information about the datasets, experimental settings, evaluation metrics, and analysis of results.

\subsection{Datasets}
Our evaluation utilizes two research collaboration network datasets, summarized in Table \ref{tab:datasets summary}.
\begin{table}[htbp]
    \centering
    \caption{Datasets Summary}
    \begin{tabular}{lcc}
        \toprule
                           & Ca-netscience & Ca-sandi-auths \\
        \midrule
        \#Nodes            & 379           & 86             \\
        \#Edges            & 914           & 124            \\
        \#AvgDegree        & 2.4           & 1.4            \\
        Directed           & TRUE          & TRUE           \\
        \%Training Edges   & 0.7           & 0.7            \\
        \%Validation Edges & 0.1           & 0.1            \\
        \%Testing Edges    & 0.2           & 0.2            \\
        \bottomrule
    \end{tabular}%
    \label{tab:datasets summary}%
\end{table}%

The Ca-netscience dataset comprises 379 nodes and 914 edges, with an average node degree of 2.4. In contrast, Ca-sandi-auths exhibits a more sparse structure with fewer nodes, edges, and a lower average node degree. Both datasets are directed networks. For experimental purposes, we employ a 70-10-20 split ratio, where 70\% of the data is used for training, 10\% for validation (early stopping), and 20\% for testing.

\subsection{Experiment Settings and Evaluation Metrics}
We benchmark FGAT against several prominent baseline models: MLP, GCN \cite{kipf2016semi}, GraphSAGE \cite{hamilton2017inductive}, and GAT \cite{velivckovic2017graph}. All baseline models maintain their default parameter configurations. For FGAT implementation, we configure the embedding dimension to 128 and employ a stack of 4 FGAT convolution layers. The dataset partitioning follows the aforementioned 0.7:0.1:0.2 ratio for training, validation, and testing, respectively.

To ensure a comprehensive performance assessment, we employ multiple evaluation metrics: Precision, Recall, F1 score, and ROC score. This diverse set of metrics provides a multifaceted evaluation, enabling thorough analysis of each model's capabilities across different performance aspects.

\subsection{Results}
The experimental results are presented in Table \ref{tab:experiment results}, yielding several significant observations:
\begin{itemize}
    \item On the Ca-netscience dataset, MLP demonstrates the poorest performance, attributable to its inability to capture spatial information encoded in the adjacency matrix. GCN, GraphSAGE, and GAT exhibit comparable performance levels, with GraphSAGE achieving superior Recall scores and GAT excelling in F1 metrics. For the Ca-sandi-auths dataset, MLP achieves notable Recall but relatively inferior Precision, suggesting overfitting tendencies and limited generalization capability. GAT, leveraging its attention mechanism, achieves the highest ROC scores among baseline methods.
    \item FGAT outperforms baseline methods across both datasets in terms of the average of four evaluation metrics. Specifically, it demonstrates an average improvement of 7.11\% across all metrics on Ca-netscience, and a more substantial 15.55\% improvement on Ca-sandi-auths. This superior performance can be attributed to two key factors: the fuzzy negative sampling mechanism, which enables focused learning on error-prone edges, and the FGAT layer architecture, which provides robust message aggregation capabilities.
\end{itemize}

\section{Conclusion}
The proposed FGAT framework, combining FNS and a novel graph attention layer, significantly improves link prediction performance compared to existing methods. FNS effectively identifies informative negative edges by leveraging fuzzy rough sets, leading to more focused and efficient model training. The FGAT layer, integrating fuzzy set concepts, captures complex relationships in graph data, resulting in superior node representations for accurate link prediction. The paper's findings highlight the potential of fuzzy rough sets in advancing GNNs for link prediction tasks and pave the way for future research exploring fuzzy set theory in graph-based learning.

\end{document}